\documentclass[runningheads]{llncs}
\usepackage[T1]{fontenc}
\usepackage{graphicx}
\usepackage{makecell}
\usepackage{subcaption}
\usepackage{diagbox}
\usepackage{multirow}
\usepackage{amsmath}
\usepackage{amssymb}

\begin{document}
\title{Efficient Logic Gate Networks \\ for Video Copy Detection}
%
%
\author{Katarzyna Fojcik\inst{1}\orcidID{0000-0002-6627-742X} 
}
\authorrunning{K. Fojcik}
%
\institute{Department of Artificial Intelligence, Wroclaw University of Science and Technology, Wroclaw, Poland}
\maketitle              
\begin{abstract}
Video copy detection requires robust similarity estimation under diverse visual distortions while operating at very large scale. Although deep neural networks achieve strong performance, their computational cost and descriptor size limit practical deployment in high-throughput systems. In this work, we propose a video copy detection framework based on differentiable Logic Gate Networks (LGNs), which replace conventional floating-point feature extractors with compact, logic-based representations.
Our approach combines aggressive frame miniaturization, binary preprocessing, and a trainable LGN embedding model that learns both logical operations and interconnections. After training, the model can be discretized into a purely Boolean circuit, enabling extremely fast and memory-efficient inference. We systematically evaluate different similarity strategies, binarization schemes, and LGN architectures across multiple dataset folds and difficulty levels.
Experimental results demonstrate that LGN-based models achieve competitive or superior accuracy and ranking performance compared to prior models, while producing descriptors several orders of magnitude smaller and delivering inference speeds exceeding 11k samples per second. These findings indicate that logic-based models offer a promising alternative for scalable and resource-efficient video copy detection.

\end{abstract}
\section{Introduction}

The rapid growth of online video platforms has intensified the need for scalable systems capable of comparing videos and identifying visual or semantic relationships. Video similarity computation underpins many tasks, including retrieval, event analysis, and copyright protection. While retrieval focuses on relative ranking, detection tasks require well-calibrated similarity scores to support threshold-based decisions, with the notion of relevance ranging from shared events to near-duplicate content.

\begin{figure}[t]
    \centering
    \begin{subfigure}[t]{0.30\textwidth}
        \centering
        \includegraphics[width=0.8\linewidth]{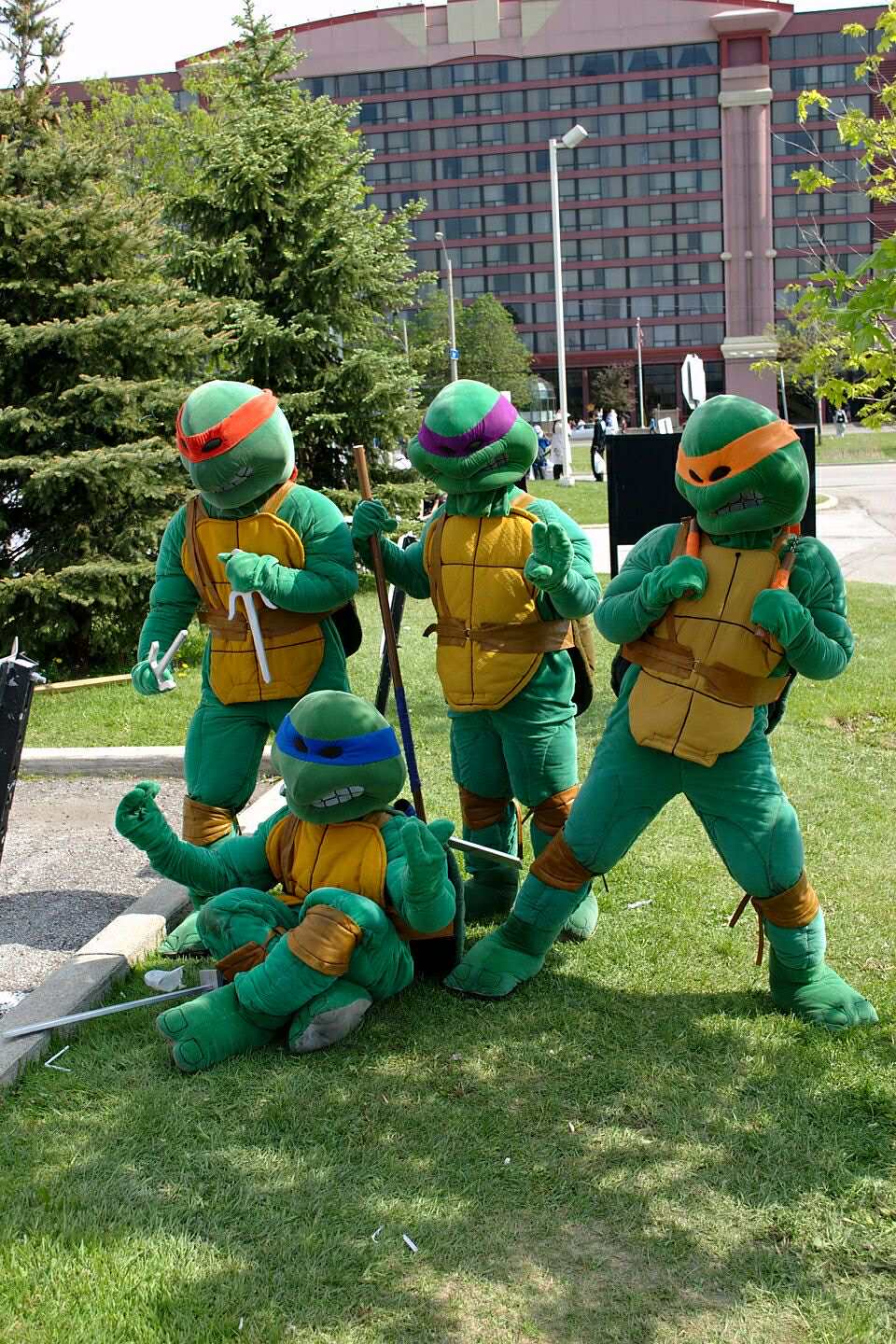}
        \caption{original}
        \label{fig:a}
    \end{subfigure}
    \hfill
    \begin{subfigure}[t]{0.30\textwidth}
        \centering
        \includegraphics[width=0.8\linewidth]{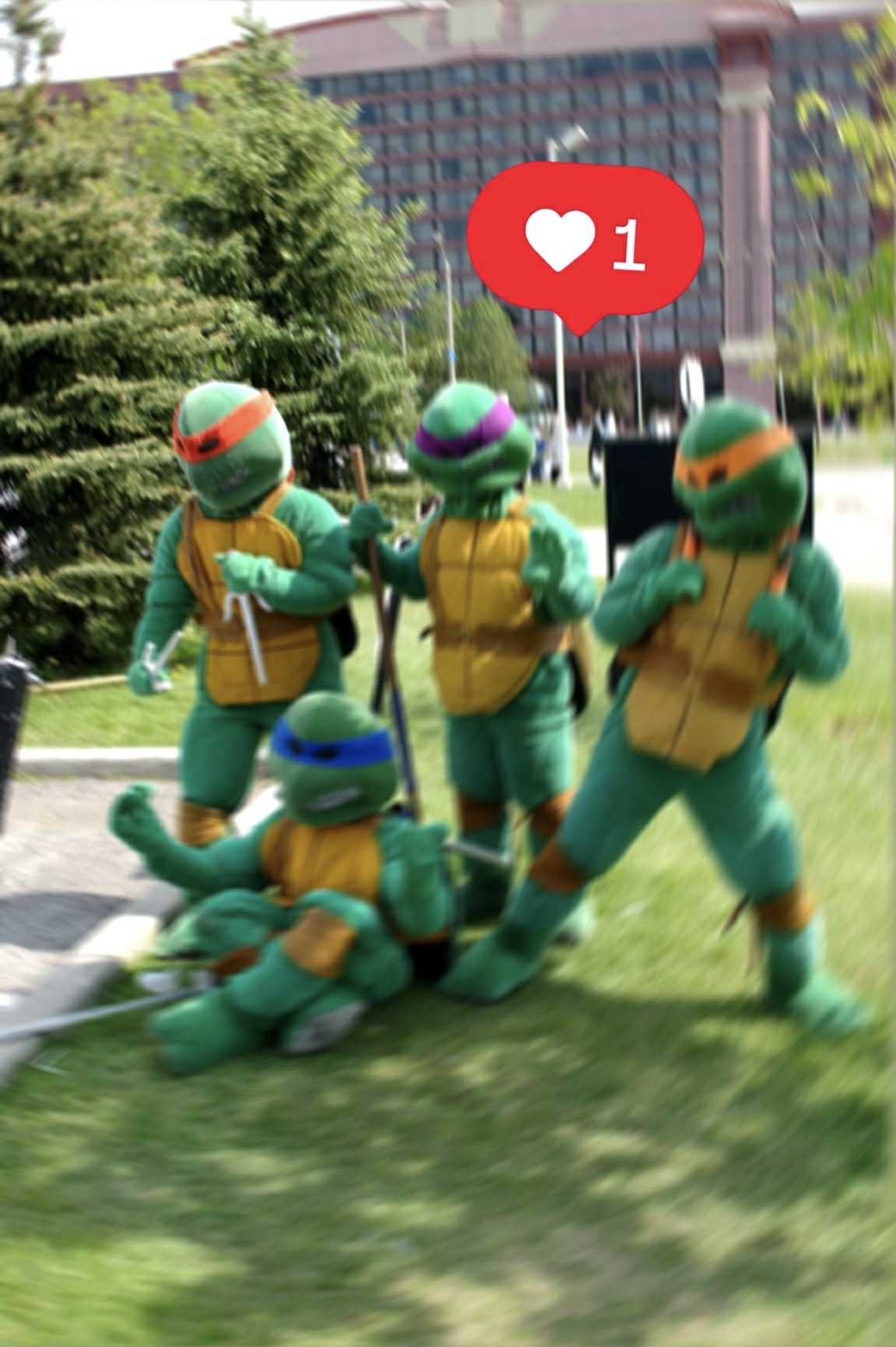}
        \caption{blur, sticker overlay}
        \label{fig:c}
    \end{subfigure}
    \hfill
    \begin{subfigure}[t]{0.30\textwidth}
        \centering
        \includegraphics[width=0.8\linewidth]{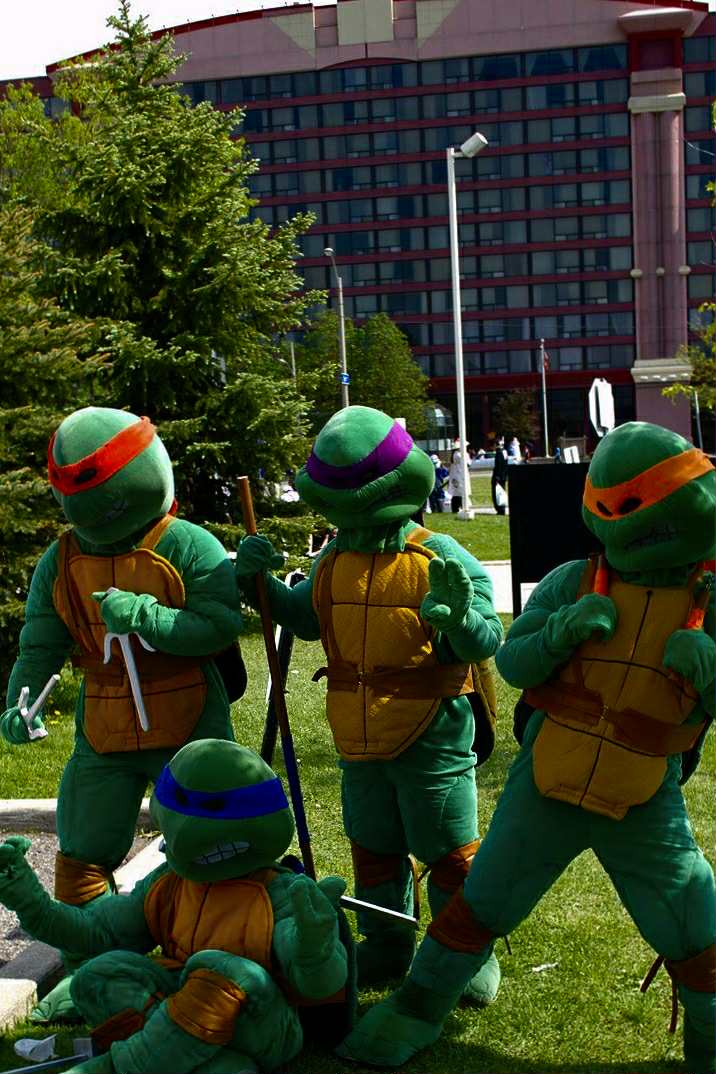}
        \caption{darken, zoom in}
        \label{fig:b}
    \end{subfigure}
    \caption{Example of video frame with applied transformations.}
    \label{fig:tranformations}
\end{figure}

Video Copy Detection (VCD) is particularly challenging, as it must identify both exact and near-duplicate videos that may undergo transformations such as re-encoding, cropping, or other augmentations (Fig.~\ref{fig:tranformations}). Large-scale platforms continuously ingest massive volumes of content, requiring VCD systems to perform reliable matching against extensive archives under strict latency constraints and in the presence of both intentional and unintentional distortions.

Although deep neural networks have substantially advanced video understanding and representation learning, their deployment in large-scale VCD pipelines remains challenging. High-performing convolutional architectures and vision transformers typically require significant computational resources, high memory bandwidth, and specialized hardware accelerators. Even optimized deep models incur inference overheads that scale poorly in high-throughput environments, motivating the search for alternative representations and inference mechanisms that preserve robustness while dramatically reducing computational cost.

\begin{figure}[t]
    \centering
    \includegraphics[width=\linewidth]{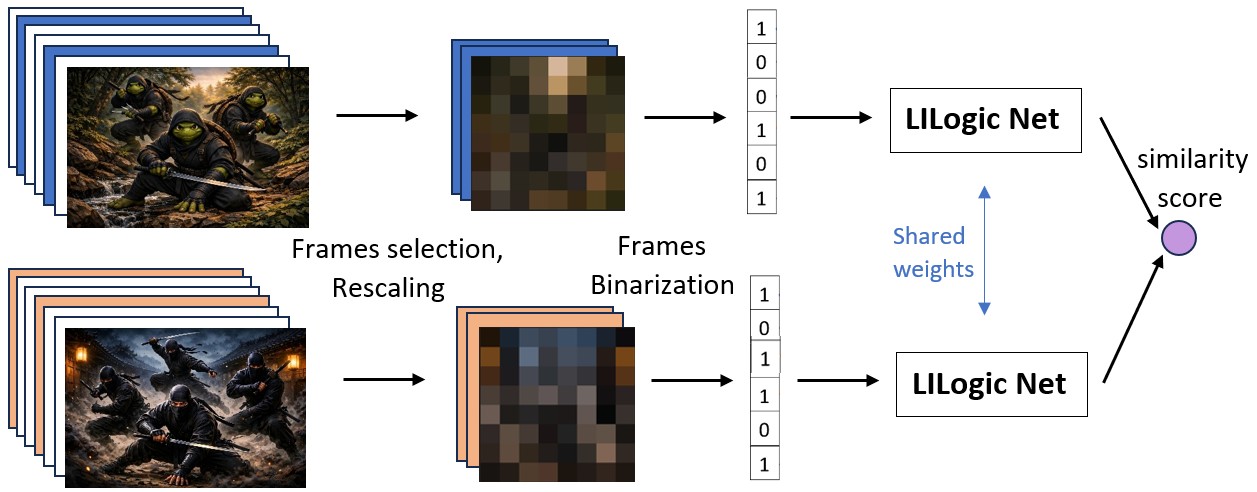}
    \caption{End-to-end workflow of the proposed framework, showing each stage from raw video preprocessing to model-based similarity computation.}
    \label{fig:schema}
\end{figure}

In prior work, we explored two directions: compact video representations using selected and aggressively miniaturized frames~\cite{fojcik2025extremely}, and LILogic Net, a differentiable logic-gate model whose learned structure can be discretized into a fixed Boolean circuit for fast inference~\cite{fojcik2025lilogic}. In this work, we unify these ideas into a single framework that replaces conventional convolutional feature extraction with a compact logic-based embedding model. We introduce the resulting pipeline (Fig.~\ref{fig:schema}), explain its conceptual foundations, and provide extensive experiments demonstrating that differentiable logic-based networks offer an effective and highly efficient alternative to conventional deep models for real-world video copy and near-duplicate detection. To the best of our knowledge, this work represents the first attempt to apply differentiable logic gate networks to a video analysis problem, and in particular to video copy and near-duplicate detection.

\section{Related work}

\subsection{Video Similarity and Copy Detection}

\subsubsection{Methods}

VCD has evolved from early handcrafted descriptors to modern deep learning approaches. Initial pipelines relied on local and global features, such as Scale-Invariant Feature Transform (SIFT) and Speeded-Up Robust Features (SURF), often combined with temporal aggregation and keyframe selection, enabling robust matching of video fragments under transformations~\cite{wary2019review}. As computational power increased, feature-based methods dominated, providing more accurate video fragment comparison.

The introduction of deep learning has fundamentally transformed VCD. Early CNN architectures, such as VGG~\cite{simonyan2014very} and ResNet~\cite{he2016deep}, provided powerful frame-level feature representations, while 3D-CNNs~\cite{li2020two} and encoder–decoder ConvLSTM models~\cite{chiang2022multi} captured temporal patterns, enhancing robustness to video modifications. More recent transformer-based models, including Video ViT~\cite{arnab2021vivit}, alongside self-supervised approaches like 3D-CSL~\cite{deng20233d}, improved the system’s ability to detect heavily edited content~\cite{deng20233d,deng2024differentiable,black2023vader}. To address efficiency and scalability, lightweight strategies such as multi-teacher distillation~\cite{ma2024let} and compact Siamese networks~\cite{fojcik2025extremely} have been developed, enabling fast and memory-efficient processing of large video collections.

Recent methods combine deep representation learning with learnable similarity functions, often leveraging attention mechanisms, recurrent networks, transformers, or frequency-domain representations to capture temporal dependencies and alignment patterns. End-to-end trainable models, such as ViSiL~\cite{kordopatis2019visil} and knowledge-distilled variants like Distill-and-Select~\cite{kordopatis2022dns} and self-supervised extensions~\cite{kordopatis2023self}, allow efficient and accurate video-to-video comparison while modeling fine-grained spatial and temporal structures.

However, these approaches often rely on complex architectures or extensive computation to achieve high accuracy. However, such approaches can be resource-intensive, limiting their practicality for large-scale video copy detection. Motivated by this, our work focuses on compact video representations combined with logic-based learning, which maintain robustness while significantly reducing memory and computational requirements.

\subsubsection{Datasets}

Several benchmark datasets have been proposed in the literature to support research on video copy detection and related problems. Datasets such as CCWeb~\cite{wu2007practical} and SVD~\cite{jiang2019svd} address the task at the collection level, providing video-level relevance annotations with respect to a given query. In contrast, FIVR-200K~\cite{kordopatis2019fivr} and EVVE~\cite{revaud2013event} employ a broader definition of video similarity by considering videos that capture the same event or incident, rather than strict copy relationships. Among the most widely used benchmarks are VCSL~\cite{he2022large} and VCDB~\cite{jiang2014vcdb}, which were specifically designed for video copy detection and video copy localization, respectively. These datasets contain partially overlapping videos and offer fine-grained, segment-level annotations that precisely indicate the temporal extent of copied content. All of the aforementioned datasets are composed of manually curated collections of user-generated videos.

In contrast to these real-world benchmarks, the META challenge dataset DVSC23~\cite{pizzi20242023} relies on synthetically generated queries produced using a carefully selected set of transformations that realistically emulate common video copy operations. This controlled generation strategy enables evaluation at a significantly larger scale than previous video copy localization datasets and facilitates systematic and reproducible benchmarking of competing methods. In our earlier work~\cite{fojcik2025extremely}, we also introduced a synthetic dataset, although based on a more limited set of augmentations, intended primarily as an initial testbed for exploring the capabilities of different frameworks. This dataset is derived from movie trailers, where frequent scene changes make copy identification more challenging than in videos with more homogeneous visual content. Moreover, all applied transformations are explicitly labeled, allowing for fine-grained analysis of robustness with respect to individual transformation types.

\subsection{Logic Gate Networks}
Parallel to these advances in VCD, a complementary line of research has explored binary or logic-based neural computations as an alternative to conventional floating-point deep models. Binarized neural networks (BNNs) \cite{hubara2016binarized,liu2020reactnet} reduce computational complexity but still rely on weighted operations. A more radical departure comes from Logic Gate Networks (LGNs), introduced by Petersen et al. \cite{petersen2022deep}, who showed that networks composed entirely of Boolean gates can be trained with differentiable relaxations. Subsequent works extended this idea with convolutional logic layers \cite{petersen2024convolutional} and learnable connectome~\cite{fojcik2025lilogic}.
Hardware-oriented research has highlighted the extremely low-latency potential of LGNs, with FPGA frameworks such as FINN \cite{umuroglu2017finn} and custom RISC-V instructions accelerating logic inference \cite{wang2025logic}. This direction emphasizes interpretability, formal verifiability, and strong alignment with digital hardware.

Despite substantial progress, no previous work has attempted to integrate logic-gate-based architectures with a full VCD pipeline. Existing VCD systems rely almost exclusively on floating-point CNN or transformer features, while logic networks have primarily been explored in classification contexts such as MNIST or CIFAR-10. The present work bridges this gap by merging our prior compact-frame representation for VCD with the recently introduced LILogic Net architecture~\cite{fojcik2025lilogic}, yielding an end-to-end Boolean video embedding model designed for large-scale deployment.

\section{Methodology}

\subsection{Model architecture}

\begin{figure}[t]
    \centering
    \begin{subfigure}[t]{0.30\textwidth}
        \centering
        \includegraphics[width=0.9\linewidth]{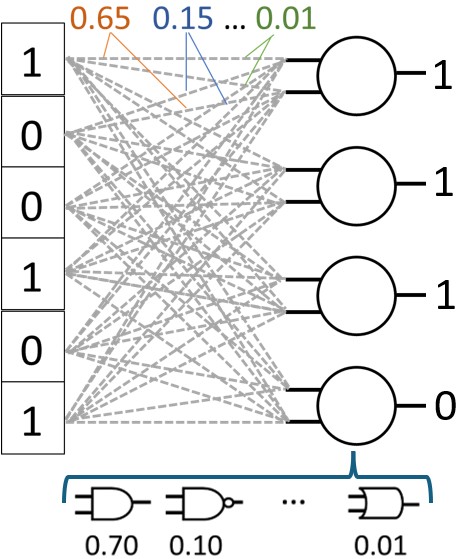}
        \caption{Dense Connectome}
        \label{fig:aa}
    \end{subfigure}
    \hfill
    \begin{subfigure}[t]{0.30\textwidth}
        \centering
        \includegraphics[width=0.9\linewidth]{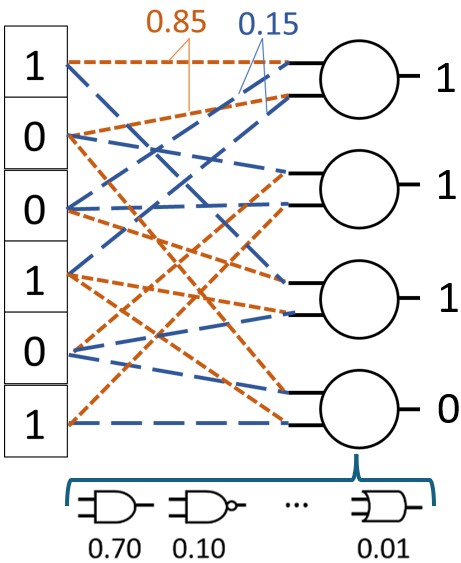}
        \caption{Sparse Connectome Top-K (here K=2)}
        \label{fig:cc}
    \end{subfigure}
    \hfill
    \begin{subfigure}[t]{0.30\textwidth}
        \centering
        \includegraphics[width=0.9\linewidth]{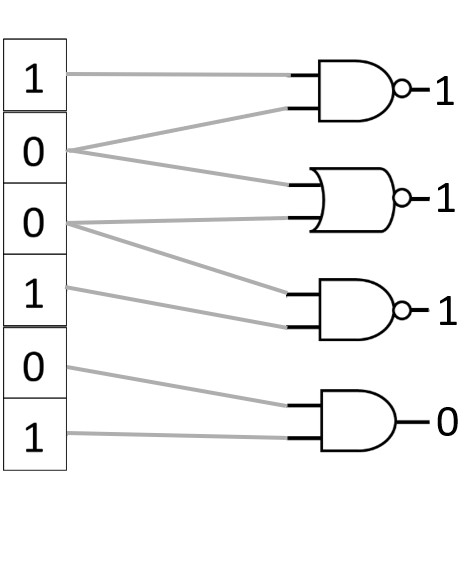}
        \caption{Fixed Connectome}
        \label{fig:bb}
    \end{subfigure}
    \caption{Overview of LILogic Net training strategies a), b) used in this study and inference pipeline c): all logic gates and connections are fully binarized.}
    \label{fig:LILogic_Net}
\end{figure}

Logic Gate Networks (LGNs) are sparse, weightless neural models composed of binary logic gates, such as AND, OR, and XOR. In their traditional form, LGNs operate on discrete inputs and implement fixed Boolean functions, which makes them extremely efficient at inference time but prevents training via gradient-based optimization. The Differentiable LGN framework introduced by Petersen et al.~\cite{petersen2022deep} addresses this limitation by relaxing the discrete nature of LGNs, enabling end-to-end training with standard optimizers.

In differentiable LGNs, binary inputs are relaxed to continuous values in the range $[0,1]$, and logic gates are interpreted probabilistically. Each gate computes a soft approximation of Boolean functions (e.g., conjunction is modeled as multiplication), and every node represents a learnable mixture over all 16 possible 2-input Boolean functions. This formulation allows gradients to propagate through both gate behavior and network structure during training. After optimization, the learned model can be discretized into a purely combinatorial LGN, yielding a stateless and highly efficient inference model.

Following the design principles of the LILogic Net framework, we employ LGN architectures that jointly learn both the gate functions and the inter-layer connectivity (connectome). This enables a systematic exploration of the trade-offs between expressiveness, sparsity, and computational efficiency.

In this study, we focus exclusively on single-layer LGN models, as prior LILogic Net results demonstrate that wide, shallow architectures provide a favorable balance between accuracy and efficiency. Each model consists of a single logic gate layer with 1,000, 2,000, 4,000 or 8,000 nodes, where each node receives exactly two inputs and implements a soft combination of the 16 Boolean functions. All nodes within a layer share the same connectome strategy.

We consider two interconnect optimization strategies (Fig.\ref{fig:LILogic_Net}):

\noindent- \textbf{Fully Learnable Dense Connectome (L)}:
For each node, both of its two inputs are connected to all outputs of the input vector. The connection weights for each input are parameterized independently using a differentiable softmax, allowing the model to learn arbitrary global connectivity patterns. While highly expressive, this variant incurs higher memory and computational cost.

\noindent- \textbf{Top-K Sparse Connectome (Top-K)}:
Each node independently selects $K$ candidate input connections for each of its two inputs at initialization. During training, a softmax is applied only over these $K$ candidates, producing a weighted combination. This enforces structured sparsity while preserving differentiability. Based on prior findings in LILogic Net, we fix $K=32$ in all Top-K experiments, which offers a strong trade-off between expressiveness and efficiency.

By restricting our evaluation to L and Top-32 variants with one-layer architectures, we directly build upon empirically validated model configurations from LILogic Net research, while enabling a controlled comparison between dense and sparse connectome learning under identical architectural constraints.

\subsection{Dataset and Preprocessing}\label{sect:Method-dataset}

We evaluate our method on a video dataset derived from movie trailers, originally introduced in prior work. The dataset is designed to assess video similarity and copy detection under challenging conditions, as trailers typically contain frequent scene changes and diverse visual content. It is based on 500 trailers collected from a publicly available YouTube playlist, with each original video accompanied by multiple distorted versions generated using perceptually meaningful transformations, such as compression, blur, noise, resizing, pixel dropout, and rotations—yielding a total of 3,000 near-duplicate pairs and 3,000 non-similar pairs. Different subsets of these distortions are used across training, validation, and testing to evaluate generalization to unseen transformations.

To obtain compact yet informative video representations, frames are selected using a content-aware strategy that favors visually distinct scenes~\cite{fojcik2025extremely,fojcik2025counteracting}. The resulting sequences are segmented into fixed-length video fragments of 15 frames, corresponding to approximately 30 seconds of video content. This fragment length provides a practical balance between robustness to partial overlaps and computational efficiency, and reflects the prevalence of short-form online video content.

\subsubsection{Frame rescaling}
Following prior work on compact CNN-based video representations, experiments on this dataset have also explored the use of low-resolution frame thumbnails. In particular, frame sizes of $32 \times 32$ and $8 \times 8$ have been considered as alternatives to standard high-resolution inputs. Although aggressive downscaling reduces per-frame visual detail, previous studies have shown that temporal sequences of such frames remain sufficiently informative for effective video copy detection.

\subsubsection{Frame binarization}
Since LGNs require binary inputs, each frame is converted into a binary representation prior to processing. The RGB channels are handled independently and binarized using fixed intensity thresholds. We evaluate two threshold sets, ${0.2, 0.4, 0.6, 0.8}$ and ${0.125, 0.25, 0.375, 0.5,}$ ${0.625, 0.75, 0.875}$, producing multiple binary feature maps per channel. These binary maps are concatenated and flattened into a single input vector per frame. This results in an effective input dimensionality of ${frame}$ ${size}^2 \times N_{\text{channels}} \times N_{\text{thresholds}}$; for example, $32 \times 32$ frames with three channels and seven thresholds yield a 21{,}504-dimensional binary input representation.

\subsection{Training and Evaluation}
All models were trained for a maximum of 300 epochs using binary cross-entropy loss, a learning rate of 0.1, and a batch size of 128. We did not apply explicit regularization methods such as dropout or weight decay, in order to focus exclusively on the impact of the LGN architecture. The Top-K interconnect variants implicitly regularize the model by limiting the connectivity space.

\setlength{\arraycolsep}{6pt}{
\begin{center}
\captionof{table}{Dataset cross validation partition; . - training, V - validation, T - test}.\label{tab:folds} 
\begin{tabular}{|c|cccccccccccc|}
  \hline
  \textbf{augmentation} & \textbf{1} & \textbf{2} & \textbf{3} & \textbf{4} & \textbf{5} & \textbf{6} & 
  \textbf{7} & \textbf{8} & \textbf{9} & \textbf{10} & \textbf{11} & \textbf{12} \\

  \hline
   compression   & V & . & . & . & . & T & . & . & . & T & V & .  \\
  
   dropout       & T & V & . & . & . & . & . & T & V & . & . & . \\
  
   gaussian blur & . & T & V & . & . & . & V & . & . & . & . & T  \\
  
   multi-noise    & . & . & T & V & . & . & . & . & T & V & . & . \\
  
   resize        & . & . & . & T & V & . & T & V & . & . & . & . \\
  
   rotation      & . & . & . & . & T & V & . & . & . & . & T & V  \\

  \hline
\end{tabular}
\end{center}
}

\subsubsection{Train–Validation–Test Splits}
Following the evaluation protocol established in prior work~\cite{fojcik2025extremely}, the dataset is split into training, validation, and test sets in a 4:1:1 ratio. The same cross-validation scheme and data folds are adopted without modification. Specifically, for each fold, one augmentation type is used for validation and another for testing, resulting in 36 predefined training–validation–test folds. In this work, we report results for 12 of these folds (Tab.~\ref{tab:folds}), ensuring that each augmentation type appears twice in both validation and testing. Using the same folds as prior work enables a fair and direct comparison while allowing us to analyze the challenges posed by different augmentation types and the model’s robustness to unseen transformations.

\subsubsection{Differentiable Relaxation}
Logic Gate Networks operate on discrete logic operators, which makes direct optimization with gradient-based methods infeasible. To overcome this limitation, we replace hard gate selection with a continuous relaxation in which each logic gate represents a soft combination of all possible two-input Boolean functions. Each gate is parameterized by a set of learnable real-valued scores that define a categorical distribution over the 16 Boolean operators. A standard softmax transformation converts these scores into non-negative coefficients that sum to one. During training, this distribution allows the model to smoothly interpolate between different logical behaviors, enabling gradients to flow through the gate selection process.

Rather than evaluating every Boolean operator explicitly, the learned gate distribution is mapped into a compact and computationally efficient representation. This is achieved through a fixed linear projection that expresses any two-input Boolean function as a combination of a small set of continuous basis terms. We use the basis ${1, A, B, A \cdot B}$, which is sufficient to represent all Boolean operators under the adopted relaxation. The projection is implemented using a fixed matrix $W_{16 \rightarrow 4}$, which encodes the exact decomposition of each Boolean operator in the chosen basis:

{
\setlength{\arraycolsep}{6pt}

\begin{equation}
W_{16 \rightarrow 4} =
\left[
\begin{array}{r|rrrr}
\text{Operator} & 1 & A & B & A \cdot B \\
\hline
\text{False} & 0 & 0 & 0 & 0 \\
A \land B & 0 & 0 & 0 & 1 \\
\lnot (A \Rightarrow B) & 0 & 1 & 0 & -1 \\
A & 0 & 1 & 0 & 0 \\
\lnot (A \Leftarrow B) & 0 & 0 & 1 & -1 \\
B & 0 & 0 & 1 & 0 \\
A \oplus B & 0 & 1 & 1 & -2 \\
A \lor B & 0 & 1 & 1 & -1 \\
\lnot (A \lor B) & 1 & -1 & -1 & 1 \\
\lnot (A \oplus B) & 1 & -1 & -1 & 2 \\
\lnot B & 1 & 0 & -1 & 0 \\
A \Leftarrow B & 1 & 0 & -1 & 1 \\
\lnot A & 1 & -1 & 0 & 0 \\
A \Rightarrow B & 1 & -1 & 0 & 1 \\
\lnot (A \land B) & 1 & 0 & 0 & -1 \\
\text{True} & 1 & 0 & 0 & 0 \\
\end{array}
\right]
\label{eq:w_matrix_logical}
\end{equation}
}

The projected coefficients determine how the gate combines its two relaxed inputs and their interaction term. This formulation preserves differentiability and interpretability while significantly reducing computational cost. After training, the soft gate representation can be collapsed to a single Boolean operator, resulting in a discrete, stateless LGN suitable for efficient inference.

\subsubsection{Learnable Interconnections}
For our architectures with adaptive connectivity, the network topology is optimized jointly with the logic gate functions. In both cases, each gate determines its two input signals using a differentiable weighting mechanism that assigns relative importance to a set of candidate connections.

Under the Top-$K$ scheme, each gate is restricted to a fixed subset of $K$ candidate input sources, selected at initialization. During training, only these candidates participate in the input selection process, with a softmax over their associated weights producing a weighted combination for each input. This approach introduces controlled sparsity into the connectome while maintaining full differentiability.

By contrast, the Fully Learnable (L) variant imposes no such restriction: each gate may connect to any output of the preceding layer (or to the raw input features in the first layer). A softmax is applied over the complete set of possible connections, allowing the model to learn highly flexible wiring patterns, albeit with increased computational and memory requirements.

\subsubsection{Similarity Computation Between Video Fragments}
Given two video fragments, frame-level embeddings are extracted using a shared frame encoder. For a batch of size $B$ and temporal length $T$, each video is represented as $\mathbf{E} \in \mathbb{R}^{B \times T \times D}$. Based on these representations, we consider three strategies to compute a single similarity score between video fragments:

    \noindent- \textbf{Concatenation-Based Similarity.} Frame embeddings are concatenated along the temporal dimension to form a $T \cdot D$-dimensional vector per fragment, and cosine similarity is computed between the resulting representations.
    
    \noindent- \textbf{Temporal Average Pooling.} Frame embeddings are averaged over time to obtain a single $D$-dimensional video-level representation, followed by cosine similarity.
    
    \noindent- \textbf{Frame-Pair Max Similarity.} All frame embeddings are $\ell_2$-normalized and a full pairwise cosine similarity matrix of size $T \times T$ is computed between fragments. The final similarity score is obtained via global max pooling over the matrix.

\noindent For all approaches, the similarity scores are linearly rescaled from $[-1,1]$ to $[0,1]$.

\subsection{Discretization}
After training, the network is transformed into a fully discrete, inference-friendly form by binarizing both the gate selections and the interconnections. This is achieved by taking the highest-probability choice (mode) from each softmax distribution. As a result, each logic gate operates on a fixed pair of binary inputs and implements a single Boolean function. At inference time, every node consistently uses two binary inputs, allowing the entire model to run purely with Boolean operations, without any floating-point computation. This discretization greatly improves both speed and memory efficiency, making the model well-suited for deployment in embedded systems or other hardware-constrained environments.

\subsection{Metrics}
Since the dataset is balanced, we report accuracy, precision, recall, and F1-score on the test set. The similarity threshold for copy/non-copy classification is selected on the validation set by maximizing the F1-score. 

In addition, we report micro Average Precision (µAP), a standard metric in video copy detection. µAP evaluates the overall ranking quality of predicted similarity scores across all query-video pairs. Specifically, all predicted similarity scores are pooled, and relevant (copy) and non-relevant (non-copy) pairs are used to compute a global precision-recall curve. µAP corresponds to the area under this curve, measuring how well the system ranks true copies above non-copies, independent of a specific threshold.

\section{Experiments}
We evaluate all LGN variants (8 model types at 2 image resolutions) across 12 dataset folds. Experiments are designed to study the impact of similarity computation, binarization thresholds, and preprocessing, as well as to assess model robustness, perform ablations, and compare with prior work. Each fold results were averaged over 5 trials.

\subsection{Comparison of Video Similarity Strategies}
To select the most effective similarity strategy, we evaluated the three considered approaches on two representative folds: fold 1 (medium difficulty) and fold 5 (hardest), based on prior work. The comparison was performed for 8×8 frames, 4 binarization thresholds, and the Top-32 LILogic Net variant with 2,000 gates (2000-Top32). Results are reported in Table~\ref{tab:similarity_comparison}. Across both folds, the frame-pair max pooling method consistently achieved the highest accuracy, F1, and $\mu$AP, demonstrating superior robustness to both moderate and challenging transformations. Accordingly, this similarity strategy was adopted for all subsequent experiments.

\subsection{Effect of Image Binarization Thresholding}

To analyze the influence of binarization granularity, we compared configurations with 4 and 7 thresholds across two image resolutions (8×8 and 32×32). This allows us to verify whether the optimal thresholding strategy depends on spatial resolution. As shown in Table~\ref{tab:threshold_resolution}, increasing the number of thresholds does not consistently improve performance. While both threshold settings yield comparable results on the easier fold, the 4-threshold configuration shows noticeably better robustness on the hardest fold, particularly in recall and F1. This trend is consistent for both image resolutions, indicating that coarser binarization generalizes better under severe distortions. Based on these observations, we adopt 4 binarization thresholds in the remaining experiments.

\begin{table}[t]
\centering
\caption{Comparison of similarity methods on fold 1 (medium) and fold 5 (hardest), evaluated using the 2000-Top32 LILogic Net model.}
\label{tab:similarity_comparison}
\begin{tabular}{|c|ccccc|ccccc|}
\hline
Method & \multicolumn{5}{c|}{Fold 1} & \multicolumn{5}{c|}{Fold 5} \\
 & Acc & Prec & Recall & F1 & $\mu$AP & Acc & Prec & Recall & F1 & $\mu$AP \\
\hline
Concat Pooling & 0.885 & 0.881 & 0.890 & 0.886 & 0.964 & 0.821 & 0.842 & 0.790 & 0.815 & 0.903 \\
Average Pooling & 0.946 & 0.908 & 0.992 & 0.948 & 0.995 & 0.871 & 0.928 & 0.804 & 0.862 & 0.955 \\
Frame-Pair Max Pool & 0.987 & 0.975 & 1.000 & 0.987 & 0.998 & 0.965 & 0.987 & 0.942 & 0.964 & 0.992 \\
\hline
\end{tabular}
\end{table}

\begin{table}[t]
\centering
\caption{Impact of binarization thresholds on similarity performance for the selected similarity method. Results are reported for fold 1 (medium difficulty) and fold 5 (hardest), evaluated using the 2000-Top32 LILogic Net model.}
\label{tab:threshold_resolution}
\begin{tabular}{|cc|ccccc|ccccc|}
\hline
Frame & Thresholds & \multicolumn{5}{|c|}{Fold 1} & \multicolumn{5}{c|}{Fold 5} \\
Size & & Acc & Prec & Recall & F1 & $\mu$AP & Acc & Prec & Recall & F1 & $\mu$AP \\
\hline
8x8  & 4 & 0.987 & 0.975 & 1.000 & 0.987 & 0.998 & 0.965 & 0.987 & 0.942 & 0.964 & 0.992 \\
8x8  & 7 & 0.987 & 0.975 & 1.000 & 0.987 & 1.000 & 0.923 & 0.986 & 0.858 & 0.918 & 0.987 \\
32x32 & 4 & 0.982 & 0.965 & 1.000 & 0.982 & 1.000 & 0.793 & 0.980 & 0.598 & 0.743 & 0.980 \\
32x32 & 7 & 0.983 & 0.967 & 1.000 & 0.983 & 1.000 & 0.770 & 0.979 & 0.552 & 0.706 & 0.980 \\
\hline
\end{tabular}
\end{table}

\begin{table}[h!]
\centering
\caption{Ablation study comparing preprocessing-only similarity against an LGN-based representation (here: 2000-Top32 LILogic Net). Results are shown for fold 1 (medium difficulty) and fold 5 (hardest).}
\label{tab:ablation_similarity}
\begin{tabular}{|c|ccccc|ccccc|}
\hline
Method & \multicolumn{5}{|c|}{Fold 1} & \multicolumn{5}{c|}{Fold 5} \\
 & Acc & Prec & Recall & F1 & $\mu$AP & Acc & Prec & Recall & F1 & $\mu$AP \\
\hline
without model & 0.891 & 0.883 & 0.902 & 0.892 & 0.879 & 0.837 & 0.886 & 0.774 & 0.826 & 0.867 \\
with 2000-Top32 model &  0.987 & 0.975 & 1.000 & 0.987 & 0.998 & 0.965 & 0.987 & 0.942 & 0.964 & 0.992 \\
\hline
\end{tabular}
\end{table}

\subsection{Ablation Study: Effect of LGN Modeling}

To isolate the impact of LGN modeling, we compare a preprocessing-only baseline against a representative LGN configuration. In the baseline setting, similarity is computed directly on resized and binarized frames, whereas the LGN-based variant additionally learns a structured logical representation. Results for a medium-difficulty fold (fold 1) and the hardest fold (fold 5) are shown in Table~\ref{tab:ablation_similarity}. While the preprocessing-only approach already achieves reasonable performance, incorporating the LGN leads to consistent and substantial improvements across all metrics, with particularly large gains in F1 and $\mu$AP. This demonstrates that the observed performance is not solely due to preprocessing, but is significantly enhanced by the learned logical structure of the LGN.

\subsection{Evaluation of LGN Model Variants Across Folds}

Table~\ref{tab:evaluation} summarizes the performance of Top-32 and Fully Learnable (L) LGN variants across all 12 folds, reporting mean values for Accuracy, F1, and $\mu$AP. Across both connectivity strategies, models operating on 8×8 frames consistently outperform their 32×32 counterparts. This suggests that coarser spatial resolution is more effective for the considered task, likely because it emphasizes global visual structure over fine-grained details that are less stable under distortions.

\begin{table}[t]
\centering
\caption{Performance metrics for Top32 and L models across different frame sizes and LGN layer widths. }
\label{tab:evaluation}
\begin{tabular}{|c|c|ccc|ccc|}
\hline
Frame Size & Layer Width & \multicolumn{3}{c|}{Top32} & \multicolumn{3}{c|}{L} \\
\cline{3-8}
 & & Acc& F1 & $\mu$AP & Acc & F1 & $\mu$AP \\
\hline
8x8 & 1000 & 0.970 & 0.967 & 0.997 & 0.978 & 0.977 & \textbf{0.997} \\
8x8 & 2000 & \textbf{0.986} & \textbf{0.985} & 0.997 & 0.975 & 0.973 & \textbf{0.997} \\
8x8 & 4000 & 0.979 & 0.978 & \textbf{0.998} & \textbf{0.985} & \textbf{0.984} & \textbf{0.997} \\
8x8 & 8000 & 0.976 & 0.973 & 0.997 & 0.978 & 0.977 & \textbf{0.997} \\
\hline
32x32 & 1000 & 0.956 & 0.948 & 0.996 & 0.970 & 0.968 & \textbf{0.997} \\
32x32 & 2000 & 0.949 & 0.938 & 0.997 & 0.970 & 0.968 & 0.996 \\
32x32 & 4000 & 0.952 & 0.943 & 0.996 & 0.978 & 0.978 & 0.996 \\
32x32 & 8000 & 0.957 & 0.950 & 0.996 & 0.979 & 0.979 & 0.996 \\
\hline
\end{tabular}
\end{table}

\begin{figure}[ht]
\centering
\includegraphics[width=0.7\textwidth]{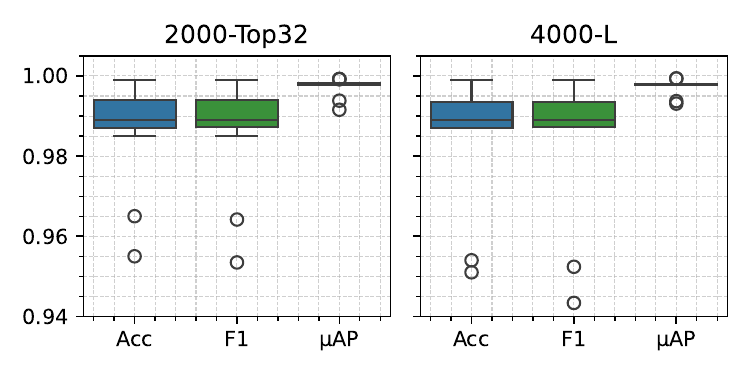}
\caption{Distribution of Accuracy, F1, and $\mu$AP across 12 evaluation folds for the 2000–Top32 and 4000–L LGN models.}\label{fig:boxplot}
\end{figure}

Among the evaluated configurations, the Top-32 model with 2000 gates and the Fully Learnable model with 4000 gates achieve the strongest overall performance, reaching median F1 scores above 0.98 and $\mu$AP values close to 1.0. These models provide an effective balance between representational capacity and robustness. The accompanying boxplots for representative configurations (2000–Top32 and 4000–L) further illustrate this stability. Accuracy and F1 distributions are tightly concentrated, indicating consistent performance across most folds, with only a small number of outliers corresponding to the most challenging test-rotation splits. In contrast, $\mu$AP exhibits very narrow distributions with minimal variance, suggesting that ranking-based performance remains largely unaffected by fold difficulty. Together, these results demonstrate that LGN models generalize well across varying conditions and that performance degradation is limited to a small subset of extreme transformations.

\subsection{Comparison with Prior Work}

\begin{table}[t]
\begin{center}
\captionof{table}{Comparison with other existing models.  The metric ``sps'' stands for samples per second, measuring the inference performance of the models. The results marked with an asterisk (*) indicate the mean results across all tested partitions, not for the entire dataset.} \label{tab:modelcomparison}
\begin{tabular}{|l|c|c|c|r|r|}  \hline
  {model} &  {accuracy}  &  {recall}  &  {precision}   &  \makecell{performance\\[0pt][sps]} &  \makecell{descriptor\\[0pt]size [kB]}     \\ \hline                                   
{ViSiL-ResNet}~\cite{kordopatis2019visil}  & {0.957}    &   {0.921}   & {\textbf{0.991}}      &  {6.5}  &  {2025} \\ \hline
{ViSiL-I3d}~\cite{kordopatis2019visil}  &  {0.778}    &  {0.792}   & {0.771}      &  {10.2}  &  {6.250} \\ \hline
{3D-CSL-base}~\cite{deng20233d}  &  {0.888}    &  {0.980}   &  {0.827}      &  {4.9}  &  {9.000} \\ \hline
{3D-CSL-small}~\cite{deng20233d}  &  {0.899}    & \ {0.919}   &  {0.884}      & {6.1}   &  {4.500} \\ \hline
{2ConvSN-8x8~\cite{fojcik2025extremely}}  &  {0842.*}    &  {0.910*}   &  {0.806*}      &  {178.6}  &  {1.875} \\ \hline
{2ConvSN-56x56~\cite{fojcik2025extremely}}  &  {0.842*}    &  {0.885*}   &  {0.823*}      &  {93.5}  &  {1.875} \\ \hline
\textbf{LILogic Net L-4000}  &  {0.985}*    &  {0.977}*   &  {0.986}*      &  {11,007.2}  &  {0.500} \\ \hline
\textbf{LILogic Net Top32-2000}  &  \textbf{0.986}*    &  \textbf{0.986}*   &  {0.985}*      &  \textbf{11,520.7}  &  \textbf{0.250} \\ \hline
\end{tabular}
\end{center}
\end{table}

Table~\ref{tab:modelcomparison} compares the proposed LGN-based models with representative deep learning approaches for video copy detection. While prior methods based on CNN and 3D architectures achieve competitive accuracy, they rely on large descriptors and incur substantial computational cost. In contrast, LILogic Net models achieve comparable or superior accuracy and recall while using extremely compact descriptors and delivering orders-of-magnitude higher inference throughput. 
Notably, the Top32-2000 and L-4000 variants achieve strong detection performance with descriptors below 1 kB and inference speeds above 11k samples/s (excluding data loading, on an NVIDIA GeForce RTX 2080 Ti), underscoring the efficiency of logic-based representations for large-scale video similarity.

\section{Conclusion}

In this work, we explored the use of logic gate networks for video copy detection, demonstrating that logic-based representations can achieve strong performance while remaining extremely compact and computationally efficient. To our knowledge, this is the first study extending LGNs from static domains, such as images or tabular data, to temporally structured video sequences.

Our experimental analysis showed that the proposed approach benefits from coarse spatial representations and structured logical modeling, leading to robust similarity estimation under common visual distortions. Across multiple evaluation folds, logic gate networks consistently improved performance compared to preprocessing-only baselines and achieved competitive results relative to existing deep learning methods, while offering orders-of-magnitude advantages in descriptor size and inference speed.

As a first step toward logic-based video analysis, this study is limited to a relatively simple dataset and a restricted set of transformations. While this setting allows for controlled evaluation, it also highlights the need for future work on more challenging benchmarks, including datasets with stronger temporal variation, complex motion patterns, and more diverse distortion types. Extending logic gate networks to such scenarios, as well as exploring deeper architectures and temporal modeling, represents an important direction for further research.

\begin{credits}
\subsubsection{\ackname} This work was supported by the Department of Artificial Intelligence at the Wrocław University of Science and Technology.

\subsubsection{\discintname}
The authors have no competing interests to declare that are
relevant to the content of this article.
\end{credits}
%
%
%
\bibliographystyle{splncs04}
\bibliography{references}
%




\end{document}